\documentclass{article}

\usepackage{arxiv}
\usepackage[utf8]{inputenc} % allow utf-8 input
\usepackage[T1]{fontenc}    % use 8-bit T1 fonts
\usepackage{hyperref}       % hyperlinks
\usepackage{url}            % simple URL typesetting
\usepackage{booktabs}       % professional-quality tables
\usepackage{amsfonts}       % blackboard math symbols
\usepackage{nicefrac}       % compact symbols for 1/2, etc.
\usepackage{microtype}      % microtypography
\usepackage{lipsum}         % Can be removed after putting your text content
\usepackage{graphicx}
\usepackage{natbib}
\usepackage{doi}
\usepackage{amsmath}
\usepackage{subcaption}     % for subfigures

\title{Sustainable Palm Tree Farming: Leveraging IoT and Multi-Modal Data for Early Detection and Mapping of Red Palm Weevil}

\author{
  \hspace{1mm}Yosra Hajjaji \\
  RIADI Laboratory, National School of Computer Science, University of Manouba, Tunisia \\
  \texttt{yossra.hajjaji@ensi-uma.tn} \\
  \And
  \hspace{1mm}Ayyub Alzahem \\
  Robotics and Internet-of-Things Laboratory, Prince Sultan University, Riyadh, Saudi Arabia\\
  \texttt{aalzahem@psu.edu.sa } \\
  \And
  \hspace{1mm}Wadii Boulila \\
  Robotics and Internet-of-Things Laboratory, Prince Sultan University, Riyadh, Saudi Arabia\\
  RIADI Laboratory, National School of Computer Science, University of Manouba, Tunisia \\
  \texttt{wboulila@psu.edu.sa} \\
  \AND
  \hspace{1mm}Imed Riadh Farah \\
  RIADI Laboratory, National School of Computer Science, University of Manouba, Tunisia \\
  \texttt{imedriadh.farah@mse.uma.tn } \\
  \And
  \hspace{1mm}Anis Koubaa \\
  Robotics and Internet-of-Things Laboratory, Prince Sultan University, Riyadh, Saudi Arabia\\
  \texttt{akoubaa@psu.edu.sa } \\
}

\begin{document}
\maketitle

\begin{abstract}
 
The Red Palm Weevil (RPW) is a highly destructive insect causing economic losses and impacting palm tree farming worldwide. This paper proposes an innovative approach for sustainable palm tree farming by utilizing advanced technologies for early detection and management of RPW. Our approach combines computer vision, deep learning (DL), the Internet of Things (IoT), and geospatial data to effectively detect and classify RPW-infested palm trees. The main phases include; (1) DL classification using sound data from IoT devices, (2) palm tree detection using YOLOv8 on UAV images, and (3) RPW mapping using geospatial data. Our custom DL model achieves 100\% precision and recall in detecting and localizing infested palm trees. The integration of geospatial data enables the creation of a comprehensive RPW distribution map for efficient monitoring and targeted management strategies. This technology-driven approach benefits agricultural authorities, farmers, and researchers in managing RPW infestations, safeguarding palm tree plantations' productivity.
\end{abstract}

% keywords can be removed

\section{Introduction}

Palm trees, with over 2,500 species and the ability to produce more than 1000 products, hold immense importance worldwide. They play a vital role in the agricultural economies of many countries, particularly in the Middle East, where they are the leading producers of dates by generating more than 8 million tons per year. Additionally, over 50\% of the world's palm oil is produced in Indonesia, which is also the leading producer \cite{ammar2011date, ecob}. The global palm oil industry is valued at around \$65 billion annually. Millions of people throughout the world have work prospects thanks to the palm tree business, especially in rural regions where palm agriculture is the main source of income \cite{9703522}. 

However, the quality and quantity of palm trees face constant risks due to various diseases, including Leaf Spots, Leaf Blights, and the notorious Red Palm Weevil (RPW) , have significant consequences \cite{ashry2022cnn, WIBOWO2022}. Due to damaged palm trees and decreased harvests, the RPW infestation alone has caused billions of dollars in economic losses worldwide. In addition, diseases like Leaf Spot and Leaf Blight can reduce palm oil production output by up to 40\%. This not only impacts farmers' livelihoods but also has negative implications for ecosystems and biodiversity. The destructive impact of the RPW has been observed in Southeast Asia, parts of Asia, northern Africa, Europe, and certain regions of North America. Human activity plays a role in facilitating the rapid spread of RPW, as diseased palm trees are inadvertently transported to non-infested areas. The RPW's ability to remain hidden inside the palm tree trunk poses challenges for early identification. If left untreated, infected palm trees can suffer long-term damage, as the RPW eventually exits the hollowed tree in search of a new host \cite{yarak2021oil, ferry2019effective, hajjaji2021improved}.
The use of Internet of Things (IoT) devices has emerged as a critical aspect of palm tree identification and management. These gadgets have a variety of sensors, including sound, temperature, and humidity sensors. The sound sensors are crucial in the context of detecting palm tree infestations. They make it possible to gather the audio data required to use deep learning (DL)-based classification techniques.

The DL algorithms can precisely recognize and classify afflicted palm trees by examining the distinctive acoustic signals generated by Red Palm Weevil (RPW) larvae or their feeding activities. Early detection of RPW infestations is essential for proactive monitoring and quick response to reduce the hazards that come with them.

This integration of DL classification based on sound data from IoT devices, as well as palm tree identification and RPW mapping phases, forms a basis of a successful method for tackling palm tree diseases and pests. This strategy provides effective and data-driven management strategies improving the general health and production of palm tree farms by utilizing the power of IoT sensors and DL algorithms. \cite{hajjaji2022leveraging, ghandorh2022semantic, hajjaji2021big}.
The value of IoT devices stems from their capacity to offer real-time data from palm tree farms, allowing for proactive monitoring, early identification, and rapid reaction to illnesses and insect infestations. The gathered sound data can be processed using DL-based classification approaches to identify individual palm tree species and assess their health state. Furthermore, combining the palm tree detection and RPW mapping phases allows for the identification and tracking of RPW-infested palms, allowing for targeted interventions to prevent and reduce the spread of this damaging pest\cite{elshafie2019red, JINTASUTTISAK2021}.

In this research, we provide a complete approach that combines deep learning health categorization based on IoT device sound data, palm tree detection algorithms, and RPW mapping methods \cite{9044348}. We want to build an effective solution for palm tree disease and pest management by integrating these components. We hope to provide precise diagnosis, early intervention, and timely actions to protect palm tree populations by analyzing sound data and applying powerful computer vision techniques.
The main contributions of our paper can be summarized as follows:
\begin{enumerate}
  \item Capturing of UAV remote sensing images and Arduino-based acoustic sensor RPW noise;
  \item Creating a dataset of Palm Tree Images Samples (PTIS) and sound Weevil data;
  \item DL palm tree health classification based on sound Weevil data and inception model;
  \item Training, validating, and testing the selected YOLOv8 models to detect and count palm trees based on PTIS;
  \item Evaluating the accuracy and performance of the optimal YOLOv8-based model for palm tree detection;
  \item Mapping the location and spatial distribution of infested and non-infested palm trees based on palm tree health classification and object detection results.
\end{enumerate}

This paper is organized as follows. Section 2 reviews relevant literature on the detection of palm tree infestation, highlighting key findings and gaps in the existing research. Section 3 describes the research design and analysis procedures used in the study with detailed explanations for every component presented in the proposed approach. Section 4, is a description of the used dataset and the experiments done on it including the results of each experiment. Finally, the main findings of the proposed study, their implications, and recommendations for future research are summarized in Section 5.

\section{Related works}

% https://www.sciencedirect.com/science/article/pii/S1110016821006967
% https://www.mdpi.com/2073-4395/10/7/987
% https://www.mdpi.com/2313-433X/8/6/170

Karar et al. \cite{karar2022smart} proposed a smart IoT-based system for detecting RPW larvae in date palms using mixed depthwise convolutional networks (MixConvNet). The system used a modified MixConvNet as a deep learning classifier to identify RPW infestation cases based on short audio recordings of feeding and/or moving RPWs. The system was tested on the public TreeVibes dataset, which contains 146 infested and 351 clean sounds. The system achieved the best accuracy score of 97.38\% and outperformed other deep learning classifiers, such as Xception and Resnet models. The system can provide an efficient and accurate solution for protecting date palm trees at the early stage of infestation.

Koubaa et al. \cite{koubaa2020smart} developed an IoT-based smart palm monitoring prototype to prevent the damage caused by the red palm weevil, a critical pest of palms. The prototype collected data from palms using smart agriculture sensors, and consumers could monitor their palm farms remotely via web/mobile applications. To interface between the sensor layer and the user layer, the authors used an industrial-level IoT platform. To examine the data and determine the infection fingerprint, they also used signal processing and statistical approaches.

Alsanea et al. \cite{alsanea2022deep} used region-based convolutional neural network (R-CNN) and convolutional neural network (CNN) techniques to detect and categorize the red palm weevil (RPW), a damaging pest of palm trees. The model located the RPW in an image using bounding boxes and retrieved characteristics to classify it. At an early stage in the Al-Qassim region, the model identified the RPW with 100\% accuracy in infested palm trees.

Ashry et al. \cite{ashry2022cnn} introduced a unique approach for the early identification of RPW, a devastating pest of palm trees using optical fiber distributed acoustic sensing (DAS) and machine learning (ML). The authors classified the signals captured by the DAS system and filtered by a 100-800 Hz bandpass filter using CNN models. In both controlled and outdoor conditions, the approach produced good accuracy and minimized false alarms, with the latter being influenced by wind noise.

The identification and monitoring of RPW have benefited greatly from these connected efforts. However, they frequently concentrated on certain elements, such as monitoring systems, image-based methods, or larvae identification. In contrast, our method offers a more complete solution for RPW identification and localization since it incorporates sound analysis, CQCC feature extraction, deep learning classification, palm tree object detection, and coordinate matching. Our method improves accuracy and allows focused intervention tactics by combining numerous data sources.

\section{Proposed approach}

\begin{figure} [ht]
\centering
\includegraphics[width=0.8\textwidth]{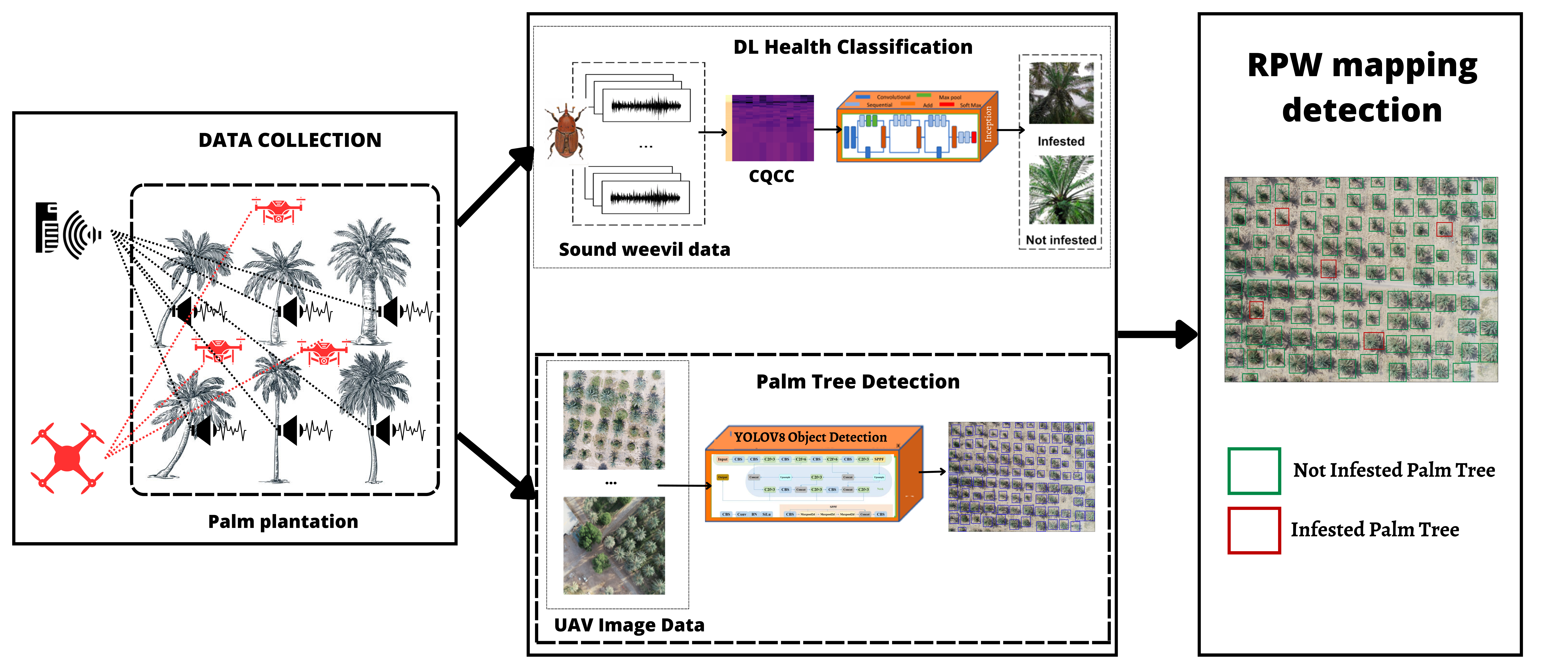}
\caption{Overview of our proposed approach of palm tree detection, counting and health classification based on integrated UAV remote sensing and sound RPW sound data.} 
\label{fig:proposed}
\end{figure}

The suggested approach's process is depicted in Figure \ref{fig:proposed} and consists of three major phases: DL classification based on sound data from IoT devices, Palm tree recognition, and RPW mapping. Each of these phases is discussed in detail in the sections that follow. The proposed methodology intends to contribute to the successful monitoring and preservation of palm tree health and productivity by integrating innovative technologies and data-driven approaches. 

\subsection{DL Health Classification}

We provide a quick method for spotting Red Palm Weevil (RPW) infestation in palm plants that makes use of reliable data and deep learning (DL) methods. The distinctive noises made by afflicted palm trees are captured using specialized equipment, which is then processed to obtain Constant-Q Cepstral Coefficients (CQCC), which stand for important characteristics of the recorded signals. The InceptionV3 model, a sophisticated DL approach, is then used to categorize and identify the existence of RPW infestation in palm trees using the retrieved data. This easy-to-use technique, though effective, permits early identification of RPW infestations and facilitates proactive steps to protect palm trees' health and wellbeing.
\subsubsection{Constant-Q Cepstral Coefficients (CQCC)}
CQCC feature extraction algorithm is a strong technique that is commonly utilized in audio signals  processing tasks such as voice recognition, sound categorization, and music analysis. CQCC captures both the spectral and temporal properties of audio signals, making it an effective representation for a wide range of audio-based applications. We present the CQCC algorithm in detail in this section, highlighting its major phases and equations.
\begin{enumerate}
    \item \textbf{Constant-Q Transform}: The first step in the CQCC algorithm is to compute the input audio signal's Constant-Q Transform (CQT). The CQT is comparable to the well-known Short-Time Fourier Transform (STFT), but with frequency bins that are logarithmically separated to better mimic human auditory perception. The following Equation defines it:
   \begin{equation}
    X[n, k] = \sum_{m=0}^{M-1} x[n+m]w[m]e^{-j2\pi kQm/M}
    \label{eq:cqcc1}
    \end{equation}
    where, $X[n, k]$ represents the CQT coefficients at time frame $n$ and frequency bin $k$. $x[n]$ denotes the input audio signal, $w[m]$ is a windowing function (e.g., Hamming window), $Q$ is the Q-factor determining the number of frequency bins per octave, and $M$ is the window length.
    \item \textbf{Log-compression and Power Normalization}: After computing the CQT coefficients, a log-compression operation is applied to enhance the perceptual representation of the audio signal. It helps to emphasize the lower magnitude components while compressing the higher ones. The log-compressed coefficients, denoted as $X_{\text{log}}[n, k]$, are obtained using the following equation:
    \begin{equation}
        X_{\text{log}}[n, k] = \log(1+\mu|X[n, k]|)
    \label{eq:cqcc2}
    \end{equation}
    where $\mu$ is a small positive constant to avoid the logarithm of zero and control the compression.

    To normalize the power across different frequency bins, a power normalization operation is performed. It aims to ensure that the resulting features are robust to variations in signal amplitude. The power-normalized coefficients, denoted as $X_{\text{norm}}[n, k]$, are computed as follows:
    \begin{equation}
        X_{\text{norm}}[n, k] = \frac{X_{\text{log}}[n, k] - \bar{X}_k}{\sigma_k}
    \label{eq:cqcc3}
    \end{equation}
    where $\bar{X}_k$ and $\sigma_k$ represent the mean and standard deviation of the CQT coefficients within each frequency bin $k$.
    \item \textbf{Cepstral Coefficients}: Next, the Cepstral Coefficients are obtained by performing the Discrete Cosine Transform (DCT) on the power-normalized coefficients. The DCT decorrelates the coefficients and retains the most salient information. The resulting Cepstral Coefficients, denoted as $C[n, k]$, are given by the equation:
    \begin{equation}
        C[n,k] = \sum_{m=0}^{M-1} X_{\text{norm}}[n, k] \cos \left( \frac{\pi k(m+0.5)}{M} \right)
    \label{eq:cqcc4}
    \end{equation}
    \item \textbf{CQCC Features}: Finally, the CQCC features are obtained by concatenating the Cepstral Coefficients across multiple time frames. This step creates a compact and discriminative representation of the audio signal suitable for further processing and analysis.
\end{enumerate}

\subsubsection{InceptionV3}
InceptionV3 is a Convolutional Neural Network (CNN) architecture developed by Google for image recognition and classification tasks. It incorporates Inception modules, and parallel convolutional layers with different filter sizes, to capture multi-scale information efficiently. InceptionV3 utilizes auxiliary classifiers for additional supervision, global average pooling for spatial information aggregation, and a softmax classifier for prediction. Trained on large-scale datasets, InceptionV3 achieves state-of-the-art performance in various computer vision tasks due to its effective multi-scale feature extraction and computational efficiency \cite{al2022thermal}.

\subsection{Palm tree detection using YOLOv8}

In this section, we aim to leverage the YOLOv8 object detection framework to achieve accurate and efficient detection of date palm trees.

YOLOv8, released in 2023 \cite{ultralytics2023yolov8}, represents a culmination of advancements from various real-time object detectors, aimed to combine the best of many real-time object detectors. It integrates the strengths of multiple models to deliver a comprehensive solution for object detection tasks, including palm tree detection. Our motivation for choosing YOLOv8 stems from its ability to leverage the best features and optimizations from different approaches, resulting in enhanced performance and accuracy. YOLOv8 incorporates state-of-the-art backbone and neck architectures (i.e., YOLOv5), improving the quality of feature extraction and enabling precise palm tree detection. The network employs a sigmoid activation function to calculate the output probability for each detection instance being a palm tree, as shown in Equation (\ref{eq1}), where $\sigma$ represents the sigmoid function and $\mathrm{net}$ denotes the input to the sigmoid.

\begin{equation}
\text{Output} = \sigma(\text{net})
\label{eq1}
\end{equation}
YOLOv8 introduces the Anchor-Free approach, departing from the Anchor-Based approach used in previous versions. Additionally, YOLOv8 incorporates the dynamic TaskAlignedAssigner for its matching strategy. To determine the alignment degree at the Anchor level for each instance, Equation \ref{eq2} is employed, where "s" represents the classification score, "u" represents the IOU value, and $\alpha$ and $\beta$ denote the weight hyperparameters. Positive samples are selected by choosing "m" anchors with the maximum value "t" in each instance, while the remaining anchors are considered negative samples. The model is then trained using the loss function. These enhancements have resulted in YOLOv8 achieving a 1\% increase in accuracy compared to YOLOv5, establishing it as the most accurate object detector to date.

\begin{equation}
t = s^{\alpha} \cdot u^{\beta}
\label{eq2}
\end{equation}

To improve the performance of the model, YOLOv8 employs a comprehensive loss function that incorporates multiple components, including classification loss, localization loss, and confidence loss, as shown in Equation (\ref{eq3}). The loss function is designed to guide the training process and optimize the detection performance, where each term contributes to different aspects of the detection task, penalizing incorrect predictions and encouraging accurate detections.
\begin{equation}
\mathrm{Loss} = \text{Classification Loss} + \text{Localization Loss} + \text{Confidence Loss}
\label{eq3}
\end{equation}

This optimized balance between accuracy and speed makes YOLOv8 an ideal choice for real-time palm tree detection in applications such as environmental monitoring, urban planning, and agriculture. Furthermore, YOLOv8 provides a variety of pre-trained models tailored to different performance requirements. These models offer a convenient starting point, leveraging the expertise and optimizations already embedded in YOLOv8. This simplifies the deployment process and saves time and resources in developing an effective palm tree detection system. Hence, YOLOv8 was chosen as the reference version as a baseline of our work \cite{lou2023dc}.

\subsubsection{Network achitecture}
YOLOv8 shares critical components/modules with YOLOv5, such as the Focus, CBL, SPP, and CSP modules. These modules serve similar purposes and functionalities in YOLOv8 as in YOLOv5. The Focus module splits the input image into four parallel slices and uses the CBL module for feature map generation. The CBL module performs convolution operations with batch normalization and leaky-ReLU activation for feature extraction. The CSP module in YOLOv8 is based on CSPNet and consists of CSP1 and CSP2 modules, similar to YOLOv5. These modules are utilized in the network backbone and neck parts of YOLOv8, where they split the input feature map, fuse cross-level features, and determine the architecture's depth. The SPP module, also present in YOLOv5, downsamples input features using max pooling layers and combines them with the starting features to capture spatial information. While the architectural elements and configurations may differ between YOLOv8 and YOLOv5, the core concepts and functions of these modules remain the same. The subsequent subsections discuss the dataset preparation, training, validation, and testing processes.

\smallskip
\subsubsection{Network training}
The training dataset included two classes: palm and tree, with accurate labeling distinguishing between palm trees and other types of trees. Rectangular bounding boxes were used to enclose the identified palm trees and trees, accounting for variations in size, overlap, occlusion, and diverse backgrounds. To enrich the training dataset, data augmentation techniques were employed, including common transformations like scaling, cropping, rotation, and color space adjustments. Mosaicking, a technique involving the combination of multiple images to create a composite image, was also used for further augmentation, introducing variations in scene composition and context. This augmented dataset significantly increased the number of training images for each class, enhancing the model's ability to generalize and accurately detect date palm trees. Various hyperparameters were adjusted to optimize the training process, including the SGD optimizer with a momentum value of 0.9 and a decay rate of 0.0005. A random seed of 45 was set for reproducibility, and the training progress was monitored with model saving every 10 epochs. Early stopping with patience of 20 epochs was implemented to prevent overfitting. These configuration settings ensured consistent and effective training of the YOLO-V8 model for accurate palm tree detection.

\subsubsection{Palm tree detection}
After training the YOLO-V8 network, we resized the test images to 1280x1280 pixels to maintain consistency in aspect ratio. This size was chosen as it struck a balance between preserving details and reducing computational load. The resized images were then fed into the trained network, generating three scales of feature maps (80x80, 40x40, and 20x20). Each scale of prediction in YOLO-V8 produced three bounding boxes per cell, containing information such as coordinates, size, confidence score, and class probability. By setting a threshold and applying Non-Maximum Suppression (NMS), low confidence boxes were removed, and redundant detections were suppressed. Finally, the bounding boxes were rescaled to the original image size to accurately locate the palm trees. This process allowed for the efficient and accurate detection of palm trees across various sizes and improved the overall performance of the system.
\subsection{RPW map generation}
In the process of generating an RPW map, the main focus is on detecting RPW infestation in palm trees and localizing their presence in UAV images. This involves a coordinated approach using recording devices attached to palm trees, UAV imagery, and palm tree object detection algorithms.

To begin, recording devices are strategically deployed and attached to individual palm trees, with the coordinates of each device pre-saved in a cloud database. These devices capture the sound of the palm trees, which are then subjected to sound analysis. By extracting CQCC features from the recorded sound, important spectral and temporal characteristics are captured.

The CQCC features are passed through a previously trained InceptionNetV3 model, which has been trained on the TreeVibes dataset. This model classifies whether a palm tree is infested or not infested with RPW based on sound analysis. The classification results are stored, along with the coordinates of each palm tree, to establish a link between the infestation status and location.

Concurrently, UAVs outfitted with cameras gather high-resolution images of the farm, focusing on the palm tree area of interest. These UAV captures show an overview of the entire farm, including the palm trees.

The next step is to apply palm tree object detection algorithms to the UAV photos, such as YOLOv8. These methods detect and localize palm tree occurrences in images, returning bounding box coordinates for each discovered palm tree.

Now comes the essential process of coordinate matching and visualization. The discovered palm tree coordinates from the object detection stage are compared to the palm tree coordinates recorded with the associated recording devices. This matching procedure aids in establishing a link between the discovered palm trees and their RPW infestation categorization.

The RPW map visualization is made once the matching is completed. The bounding boxes of the identified palm trees are colored dependent on their RPW infestation condition. If a palm tree is afflicted with RPW, the appropriate bounding box is highlighted in red, showing the existence of the infestation. If a palm tree is identified as not infected, the bounding box is depicted in blue, indicating a healthy palm tree.

The RPW map is created by integrating the findings of sound analysis, RPW classification, UAV images, and object recognition. This map depicts the spread of RPW-infested palm trees on the farm, allowing for efficient monitoring, targeted intervention, and effective RPW infestation control.

\section{Experimentation and results}

\subsection{Dataset description}
The experiments of our approach are carried out on a multi-modal dataset composed of RPW sound data and UAV image data.

\subsubsection{RPW sound data }
In this paper, we utilize the TreeVibes dataset \cite{rigakis2021treevibes}, consisting of over 54,000 10-second audio clips recorded from different types of trees in various locations and seasons. The dataset provides labels indicating tree species, location, season, weather, and time of day. It includes a wide range of tree species across continents, capturing variations in tree sounds across seasons and weather conditions. Each audio clip is accompanied by metadata, such as GPS coordinates, date and time, temperature, humidity, and air pressure. The dataset's richness is summarized in Table \ref{tab:treevibes}.

\begin{table}[h]
    \centering
    \begin{tabular}{p{3cm}|p{3cm}}
        \hline
        \textbf{Label} & \textbf{Number of records} \\ \hline
        Not infested & 967 \\ \hline
        Infested & 53676 \\ \hline
    \end{tabular}
    \caption{Number of records in TreeVibes dataset }
    \label{tab:treevibes}
\end{table}

\subsubsection{UAV image data}
We have also collected a dataset consisting of aerial images from two different locations, including a total of 349 images. The images were captured using two different drones. The first drone used was a DJI Phantom 4 Pro equipped with a DJI FC6310 camera, which has resolutions of 4864x3648 and 4096x2160. The second drone used was a DJI Mavic Pro equipped with a DJI FC220 camera, with a resolution of 4000x3000. To analyze and classify the images, the dataset was manually labeled using the Labelbox platform. A total of 13,071 instances were labeled, with 11,150 instances identified as palm trees and 1,921 instances labeled as other trees.
\subsection{Results}
\subsubsection{DL classification based on sound data}
We used a dataset of red palm weevil audio recordings, converting them into images using CQCC feature extraction. The image dataset was split into training, validation, and testing sets with ratios of 0.8, 0.1, and 0.1, respectively. InceptionV3 was employed as the deep learning model, trained with a batch size of 16, a learning rate of 0.0001, and 200 iterations. Adam optimizer and binary cross-entropy loss function were utilized.

The trained InceptionV3 model was evaluated on the testing dataset, showing highly promising results. It achieved perfect performance with an accuracy, precision, recall, and F1-score of 1.000 across all metrics.

Validation accuracy and loss were visualized at different iterations. The plots showed a steady improvement in accuracy and a decreasing trend in the loss function throughout the training process.
\begin{figure}[h]
  \centering
  \begin{subfigure}{0.45\textwidth}
    \includegraphics[width=\linewidth]{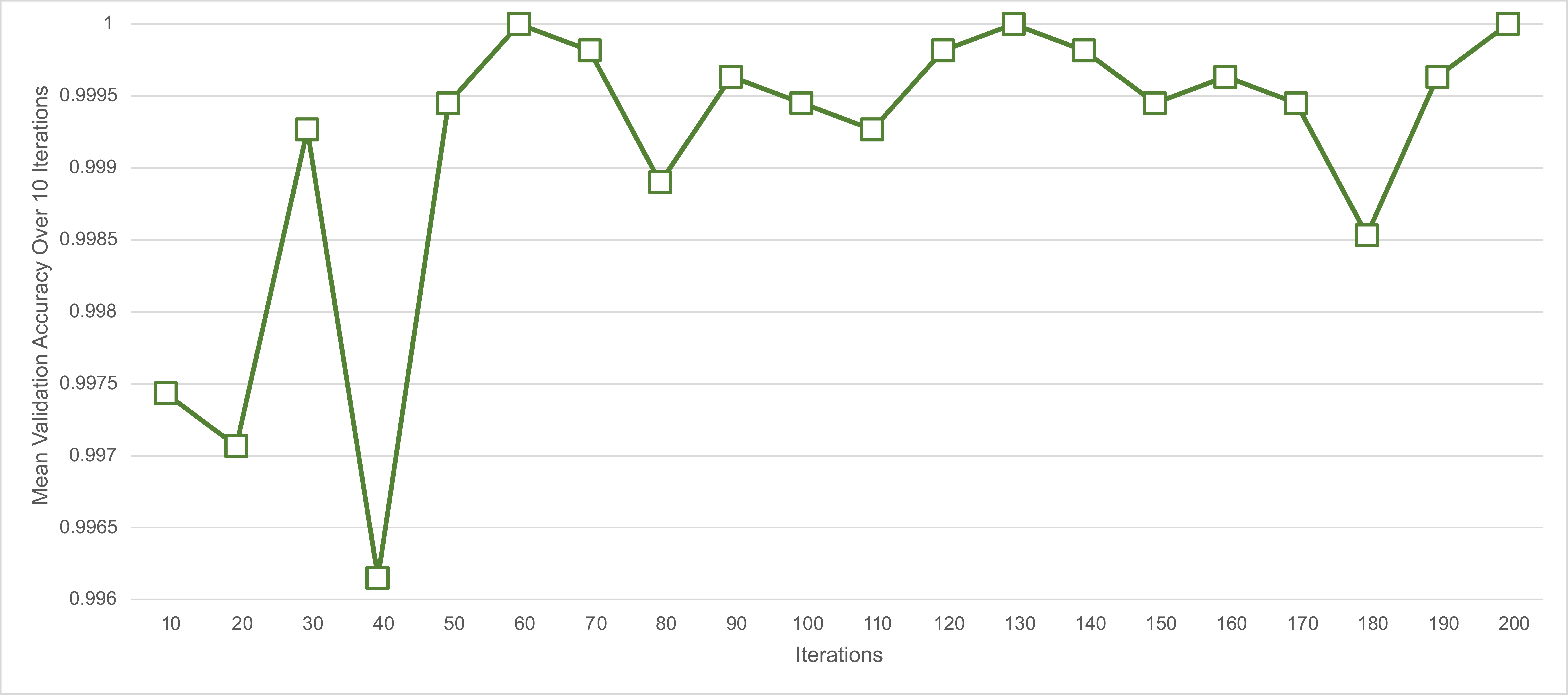}
    \caption{Mean validation accuracy over 10 iterations.}
    \label{fig:acc}
  \end{subfigure}
  \hfill
  \begin{subfigure}{0.45\textwidth}
    \includegraphics[width=\linewidth]{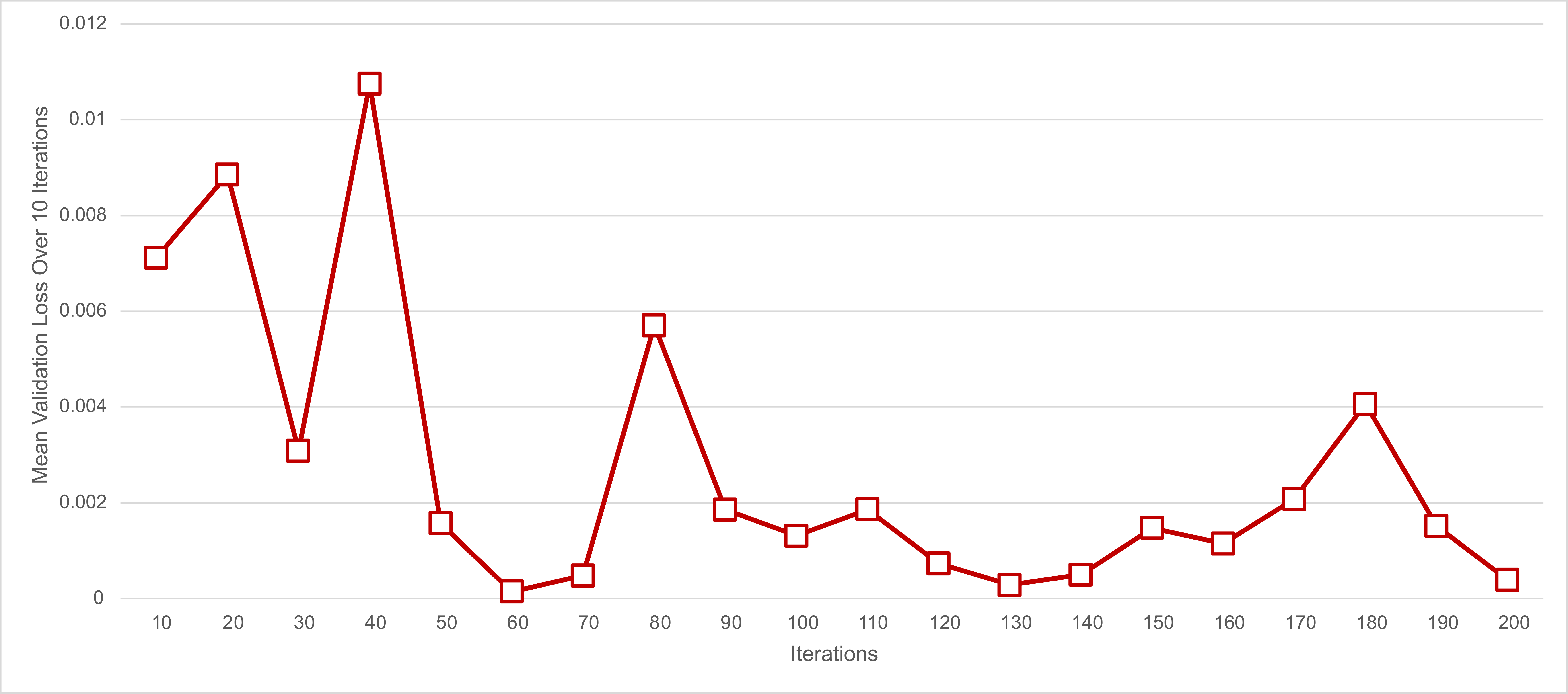}
    \caption{Mean validation loss over 10 iterations.}
    \label{fig:loss}
  \end{subfigure}
  \caption{Comparison of accuracy and loss.}
  \label{fig:comparison}
\end{figure}

\subsubsection{YoloV8 for object detection}
We trained the YOLOv8 object detection model on a carefully curated dataset for palm tree detection. Our goal was to create a precise and robust model capable of identifying palm trees in various contexts. The YOLOv8 large model architecture was used with 60 epochs of training and an image size of 1280. 
Evaluation results showed the effectiveness of our trained YOLOv8 model for palm tree detection. The precision achieved was 0.841, correctly identifying palm trees in 84.1\% of instances. The recall value reached 0.865, successfully detecting 86.5\% of actual palm trees in the dataset. To assess overall accuracy, we used Mean Average Precision (mAP) metrics. The mAP@50 value was 0.899, indicating high accuracy with a 50\% confidence threshold. The mAP@50-95 value was 0.541, showing decreased performance with a wider range of confidence thresholds.

\begin{figure}[h]
\centering
\includegraphics[width=0.8\textwidth]{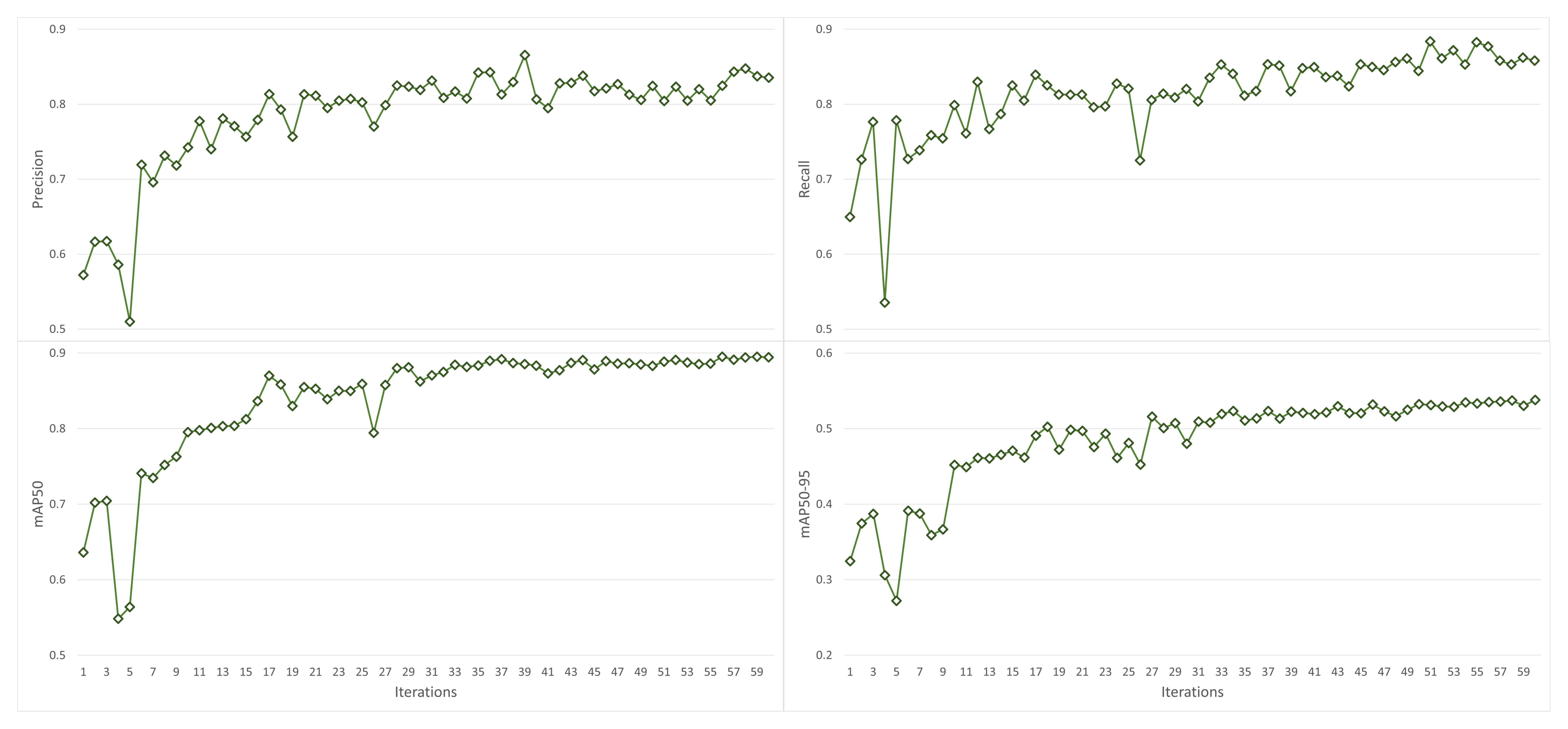}
\caption{Precision, Recall, mAP50, and mAp50-95 history.} 
\label{fig:yolo_results}
\end{figure}

Figure \ref{fig:yolo_results} presents the training progress for the palm trees object detection matrices.

\subsubsection{RPW map generation}
\begin{figure}[h]
\centering
\includegraphics[width=4in]{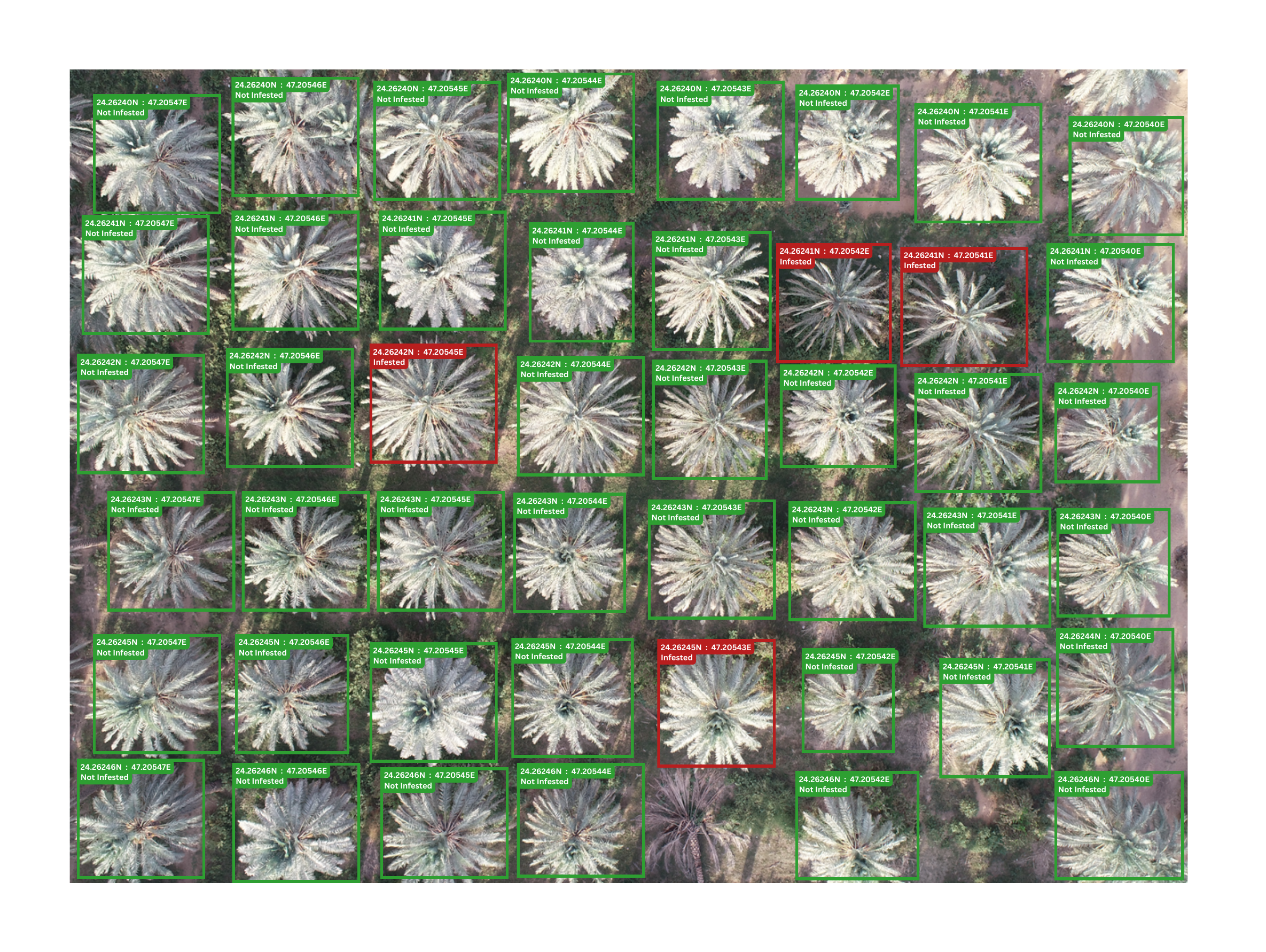}
\caption{RPW Map Visualizing Infested and Not Infested Palm Trees.} 
\label{fig:map}
\end{figure}

The RPW map generation process has yielded promising results in detecting and visualizing RPW infestation in palm trees on the farm. The results provide valuable insights into the distribution of RPW-infested palm trees, aiding in effective management strategies. Figure \ref{fig:map} showcases the RPW map, displaying the UAV image with bounding boxes representing infested and not infested palm trees.

\section{Conclusion}
In this study, our proposed approach utilizing IoT and multi-modal data offers a promising solution for the early detection and mapping of Red Palm Weevil (RPW) infestations in palm trees. By converting sound data into images using the CQCC feature and employing the InceptionV3 model for classification, we achieved 100\% accuracy in RPW detection, a significant milestone in RPW management. The integration of the YOLOv8 model for palm tree detection in UAV images and RPW mapping further enhances the effectiveness of our approach. The generated RPW distribution map provides valuable insights for farm managers to take targeted actions and protect the health and productivity of palm tree ecosystems. 
However, implementing the proposed approach on a larger scale, particularly in industrial palm tree farms, brings both advantages and challenges. Geographic variations, such as differences in vegetation density or topography, can affect the accuracy of palm tree recognition in UAV images. Therefore, these factors need to be carefully considered throughout the system's development and deployment. While the adoption of the suggested strategy may entail initial financial investments for IoT devices, sensors, and data infrastructure, it can lead to significant cost savings by enabling early identification and mapping of RPW infestations. Prompt intervention and targeted actions based on accurate information not only reduce economic losses caused by RPW infestations but also protect the productivity of palm tree ecosystems. Conducting cost-effectiveness studies considering long-term benefits and potential yield losses is essential to determine the economic feasibility of adopting this approach. The scalability of IoT devices and sensors facilitates comprehensive monitoring in large plantation areas, but managing the substantial volume of generated data and ensuring the reliability and maintenance of the IoT infrastructure pose challenges. Efficient data processing, analysis, and storage infrastructure, along with high-performance computing systems, are necessary to support the integration of deep learning algorithms. Future research should focus on validating and generalizing the proposed approach across different regions and palm tree species, while exploring the integration of additional sensing technologies such as thermal imaging or multi-spectral analysis, may provide complementary information for improved RPW detection and mapping.

\bibliographystyle{unsrtnat}
\bibliography{references}

\end{document}